\documentclass[sigconf]{acmart}

\usepackage{booktabs} 
\usepackage{tabularx} 
\usepackage{enumerate}
\usepackage{subfigure}
\usepackage{graphicx}
\usepackage{amsmath}
\usepackage{tablefootnote}
\usepackage{array,colortbl,xcolor}
\newcolumntype{Y}{>{\raggedright\arraybackslash}X} 
\renewcommand\footnotetextcopyrightpermission[1]{}
\setcopyright{none}
\makeatletter
\def\highlight#1{%
\fboxrule2pt %
\hsize=\dimexpr\hsize-2\fboxrule-2\fboxsep\relax
#1%
\@endpbox\unskip\setbox0\lastbox\bgroup
\fboxrule2pt %
\fcolorbox{red}{lightgray}{\box0}\hfill}
\acmConference[SIGKDD]{2nd ML for PHM Workshop at Special Interest Group on Knowledge Discovery and Data Mining}{Aug 2017}{Canada} 
\acmYear{2017}
\copyrightyear{2017}

\begin{document}
\title{Predicting Remaining Useful Life using Time Series Embeddings based on Recurrent Neural Networks\footnote{Copyright $\copyright$ 2017 Tata Consultancy Services Ltd.} }
\author{Narendhar Gugulothu, Vishnu TV, Pankaj Malhotra, \\ Lovekesh Vig, Puneet Agarwal, Gautam Shroff}
\affiliation{%
  \institution{TCS Research, New Delhi, India}}
\email{{narendhar.g, vishnu.tv, malhotra.pankaj,lovekesh.vig, puneet.a, gautam.shroff}@tcs.com} 
\begin{abstract}
We consider the problem of estimating the remaining useful life (RUL) of a system or a machine from sensor data. Many approaches for RUL estimation based on sensor data make assumptions about how machines degrade. Additionally, sensor data from machines is noisy and often suffers from missing values in many practical settings. We propose Embed-RUL: a novel approach for RUL estimation from sensor data that does not rely on any degradation-trend
assumptions, is robust to noise, and handles missing values. Embed-RUL utilizes a sequence-to-sequence model based on Recurrent Neural Networks (RNNs) to generate embeddings for multivariate time series subsequences. The embeddings for normal and degraded machines tend to be different, and are
therefore found to be useful for RUL estimation. We show that the embeddings capture the overall pattern in the time series while filtering out the noise, so that the embeddings of two machines with similar operational behavior are close to each other, even when their sensor readings 
have significant and varying levels of noise content. We perform experiments on publicly available turbofan engine dataset and a proprietary real-world dataset, and demonstrate that Embed-RUL outperforms the previously reported \cite{malhotra2016multi} state-of-the-art on several metrics.
\end{abstract}

\maketitle
\section{Introduction}

It is quite common in the current era of the `Industrial Internet of Things' \cite{da2014internet} 
for a large number of sensors to be  installed for monitoring the operational behavior of machines. 
Consequently, there is considerable interest in exploiting data from such sensors for health monitoring tasks such as anomaly detection, fault detection, 
as well as prognostics, i.e., estimating remaining useful life (RUL) of machines in the field. 

We highlight some of the practical challenges in using data-driven approaches for health monitoring and RUL estimation, and propose an approach that can handle these challenges:
\newline
1) \textit{Health degradation trend}: In complex machines with several components, it is difficult to build physics based models for health degradation analysis. Many data-driven approaches assume a degradation trend, e.g., exponential degradation \cite{croarkin2006nist,saxena2008damage,ramasso2014investigating,
camci2016comparison,wang2008similarity}. This is particularly useful in cases where there is no explicit measurable parameter of the health of a machine. Such an assumption may not hold in other scenarios, e.g., when a component in a machine is approaching failure, the symptoms in the sensor data may initially be intermittent and then grow over time in a non-exponential manner.
\newline 
2) \textit{Noisy sensor readings}: Sensor readings often suffer from varying levels of environmental noise which entails the use of denoising techniques. The amount of noise may even vary across sensors.
\newline
3) \textit{Partial unavailability of sensor data}: Sensor data may be partially unavailable due to several reasons such as network communication loss and damaged or faulty sensors. 
\newline
4) \textit{Complex temporal dependencies between sensors}: Multiple components interact with each other in a complex way leading to complex dependencies between sensor readings.
For example, a change in one sensor may lead to a change in another sensor after a delay of few seconds or even hours. 
It is desirable to have an approach that can capture the complex operational behavior of machine(s) from sensor readings while accounting for temporal dependencies.
\newline

In this paper, we propose \textit{Embed-RUL}: an approach for RUL estimation using Recurrent Neural Networks (RNNs) to address the above challenges. 
An RNN is used as an encoder to obtain a fixed-dimensional representation that serves as an  embedding for multi-sensor time series data. 
The health of a machine at any point of time can be estimated by comparing an embedding computed using recent sensor history with representative embeddings computed for periods of normal behavior.
Our approach for RUL estimation does not rely on degradation trend assumptions, can handle noise and missing values, and can capture complex temporal dependencies among the sensors. 
The key contributions of this work are:
\begin{itemize}
\item We show that time series embeddings or representations obtained using an RNN Encoder are useful for RUL estimation (refer Section~\ref{sec:rul_eval}).
\item We show that embeddings are robust and perform well for the RUL estimation task even under noisy conditions, i.e., when sensor readings are noisy (refer Section~\ref{sec:noise_test}).
\item Our approach compares favorably to previous benchmarks for RUL estimation \cite{malhotra2016multi} on the turbofan engine dataset \cite{saxena2008turbofan} as well as on a real-world pump dataset (refer Section~\ref{sec:rul_eval}).
\end{itemize}

The rest of the paper is organized as follows: We provide a review of related work in Section~\ref{sec:related_work}. Section~\ref{sec:background} motivates our approach and briefly introduces existing RNN-based approaches for machine health monitoring and RUL estimation using sensor data. In Section~\ref{sec:Embed-RUL} we explain our proposed approach for RUL estimation, and provide experimental details and observations in Section~\ref{sec:results}, and conclude in Section~\ref{sec:discussion}.
\section{Related Work}\label{sec:related_work}
\textit{Data-driven approaches for RUL estimation}:
Several approaches for RUL estimation based on sensor data have been proposed. 
A review of these approaches can be found in \cite{si2011remaining}.
\citep{eker2014similarity,khelif2014rul} propose estimating RUL directly by calculating the similarity between the sensors without deriving any health estimates.
Similarly, Support Vector Regression \cite{khelif2017direct}, RNNs \cite{heimes2008recurrent}, Deep Convolutional Neural Networks \cite{babu2016deep} have been proposed to estimate the RUL directly by modeling the relations among the sensors without estimating the health of the machines. 
However, unlike Embed-RUL, none of these approaches focus on robust RUL estimation, and in particular, on robustness to noise. 

\textit{Robust RUL Estimation}: 
Wavelet filters have been proposed to handle noise for robust performance degradation assessment in \cite{qiu2003robust}. In \cite{hu2012ensemble}, ensemble of models is used to ensure that predictions are robust.
Our proposed approach handles noise in sensor readings by learning robust representations from sensor data via RNN Encoder-Decoder (RNN-ED) models.

\textit{Time series representation learning}: Unsupervised representation learning for sequences using RNNs has been proposed for applications in various domains including text, video, speech, and time series (e.g., sensor data). Long Short Term Memory (LSTM) \cite{hochreiter1997long} based encoders trained using encoder-decoder framework have been proposed to learn representations of video sequences \citep{srivastava2015unsupervised}. 
Pre-trained LSTM Encoder based on autoencoders are used to initialize networks for classification tasks and are shown to achieve improved performance \citep{dai2015semi} for text applications. Gated Recurrent Units (GRUs) \cite{cho2014learning} based encoder named Timenet \citep{malhotra2017timenet} has been recently proposed to obtain embeddings for time series from several domains. The embeddings are shown to be effective for time series classification tasks. 
Stacked denoising autoencoders have been used to learn hierarchical features from sensor data in \cite{yan2015accurate}. These features are shown to be useful for anomaly detection.
However, to the best of our knowledge, the proposed Embed-RUL is the first attempt at using RNN-based embeddings of multivariate sensor data for machine health monitoring, and more specifically, for RUL estimation. 


\textit{Other Deep learning models for Machine Health Monitoring}:
Various architectures based on Restricted Boltzmann Machines, RNNs (discussed in Section~\ref{sec:rnn_health_monitoring}) and Convolutional Neural Networks have been proposed for machine health monitoring in different contexts.
Many of these architectures and applications of deep learning to machine health monitoring have been surveyed in \citep{zhao2016deep}. 
An end-to-end convolutional selective autoencoder for early detection and monitoring of combustion instabilites in high speed flame video frames was proposed in \citep{DBLP:journals/corr/AkintayoLSS16}. A combination of deep learning and survival analysis for asset health management has been proposed in \citep{kdd2016_2} using sequential data by stacking a LSTM layer, a feed forward layer and a survival model layer to arrive at the asset failure probability. 
Deep belief networks and autoencoders have been used for health monitoring of aerospace and building systems in \citep{kdd2016_3}. 
However, none of these approaches are proposed for RUL estimation.
Predicting milling machine tool wear from sensor data has been proposed using deep LSTM networks in \citep{zhao2016machine}. 
In \citep{zhao2017learning}, a convolutional bidirectional LSTM network along with fully connected layers at the top is shown to predict tool wear. The convolutional layer extracts robust local features while LSTM layer encodes temporal information. These methods model the problem of degradation estimation in a supervised manner unlike our approach of estimating machine health using embeddings generated using seq2seq models.
\section{Background}\label{sec:background}
\begin{figure}[h]
\includegraphics[trim={4.5cm 1.0cm 1.5cm 2cm},scale=0.15,clip]{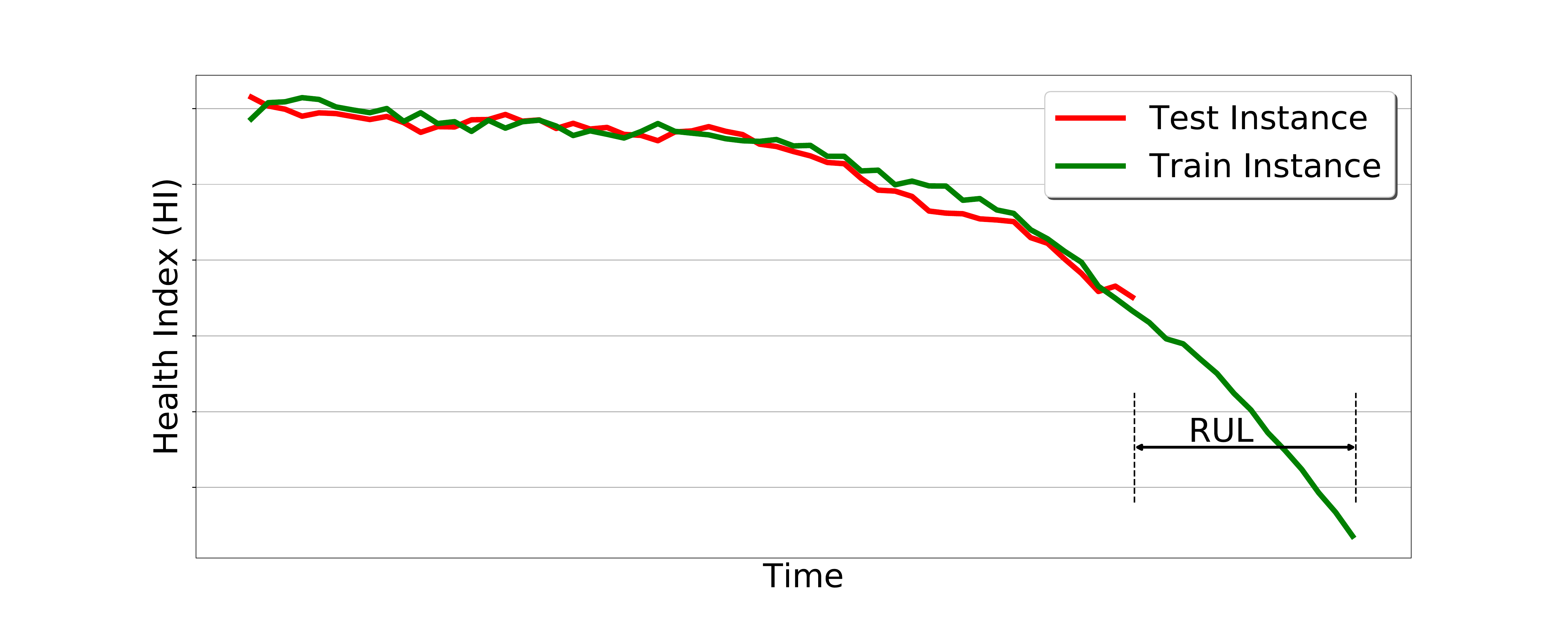}
\caption{\label{fig:Curve_matching}Example of RUL estimation using curve matching.} 
\end{figure}

Many data-driven approaches attempt to estimate the health of a machine from sensor data in terms of a \textit{health index (HI)} (e.g., \cite{wang2008similarity,ramasso2014investigating}). 
The trend of HI over time, referred to as HI curve, is then used to estimate the RUL by comparing it with the trends of failed instances.
The HI curve for a test instance is compared with the HI curve of failed (train) instance to estimate the RUL of the test instance, as shown in Figure~\ref{fig:Curve_matching}. 
In general, the HI curve of the test instance is compared with HI curves of several failed instances, and weighted average of the obtained RUL estimates from the failed instances is  used to obtain the final RUL estimate (refer Section~\ref{sec:RUL_estimation} for details).

In Section~\ref{sec:lin_reg_hi}, we introduce a simple approach for HI estimation that maps the current sensor readings to HI. Next, we introduce existing HI estimation techniques that leverage RNNs to capture the temporal patterns in sensor readings, and provide a motivation for our approach in Section~\ref{sec:rnn_health_monitoring}.

\subsection{Degradation trend assumption based HI estimation}\label{sec:lin_reg_hi} 
Consider a HI curve $H=[h_1,h_2,\ldots,h_T]$, where $0\,{\leq}\,h_t\,{\leq}\,1$, $t=1,2,\ldots,T$. 
When a machine is healthy, $h_t=1$ and when a machine is near failure or about to fail, $h_t=0$.
The multi-sensor readings $\mathbf{x}_t\in \mathbb{R}^n$ at time $t$ can be used to obtain an estimate ${h}'_t$ for the actual HI value ${h}_t$. 
One way of obtaining this mapping is via a linear regression model: 
${h}'_t=f_{\boldsymbol{\theta}}(\mathbf{x}_t)=\boldsymbol{\theta}^T\mathbf{x}_t+\theta_0$,
where $\boldsymbol{\theta}\in\mathbb{R}^{n}$ and $\theta_0\in\mathbb{R}$. 
The parameters $\boldsymbol{\theta}$ and $\theta_0$ are estimated by minimizing $\sum^{T}_{t=1}{({h}'_t-h_t)^2}$, where the \textit{target HI curve} can be assumed to follow an exponential degradation trend (e.g., \cite{wang2008similarity}).

Once the mapping is learned, the sensor readings at a time instant can be used to obtain HI. Such approaches have two shortcomings: i) they rely on an assumption about the degradation trend, ii) they do not take into account the temporal aspect of the sensor data. We show that the target HI curve for learning such a mapping (i.e., learning the parameters $\boldsymbol{\theta}$ and ${\theta}_0$) can be obtained using RNN models instead of relying on the exponential assumption (refer Section~\ref{sec:results} for details).
\subsection{RNNs for Machine Health Monitoring}\label{sec:rnn_health_monitoring}
RNNs, especially those based on LSTM units or GRUs have been successfully used to achieve state-of-the-art results on sequence modeling tasks such as machine translation \cite{cho2014learning} and speech recognition \cite{graves2013speech}. 
Recently, deep RNNs have been shown to be useful for health monitoring from multi-sensor time series data \cite{p:lstm-ad,p:icmlLSTM-AD,filonov2016multivariate}.  
The key idea behind using RNNs for health monitoring is to learn a temporal model of the system by capturing the complex temporal as well as instantaneous dependencies between sensor readings. 

Autoencoders have been used to discover interesting structures in the data by means of regularization such as by adding constraints on the number of hidden units of the autoencoder \cite{ng2011sparse}, or by adding noise to the input and training the network to reconstruct a denoised version of the input \cite{vincent2008extracting}. 
The key idea behind such autoencoders is that the hidden representation obtained for an input retains the underlying important pattern(s) in the input and ignores the noise component.

RNN autoencoders have been shown to be useful for RUL estimation \cite{malhotra2016multi} in which the RNN-based model learns to capture the behavior of a machine by learning to reconstruct multivariate time series corresponding to normal behavior in an unsupervised manner. 
Since the network is trained only on time series corresponding to normal behavior, it is expected
to reconstruct the normal behavior well and perform poorly while reconstructing the abnormal behavior. 
This results in small reconstruction error for normal time series and large reconstruction error for abnormal time series.
The reconstruction error is then used as a proxy to estimate the health or degree of degradation, and in turn estimate the RUL of the machine.
We refer to this reconstruction error based approach for RUL estimation as \textit{Recon-RUL}.

We propose to learn robust fixed-dimensional representations for multi-sensor time series data via sequence-to-sequence \cite{sutskever2014sequence,bahdanau2014neural} autoencoders based on RNNs.
Here we briefly introduce multilayered RNNs based on GRUs that serve as building blocks of sequence-to-sequence autoencoders (refer Section~\ref{sec:Embed-RUL} for details).
\begin{figure*}[h]
{\includegraphics[trim={0.0cm 2.5cm 0.0cm 1cm},scale=0.27,clip]{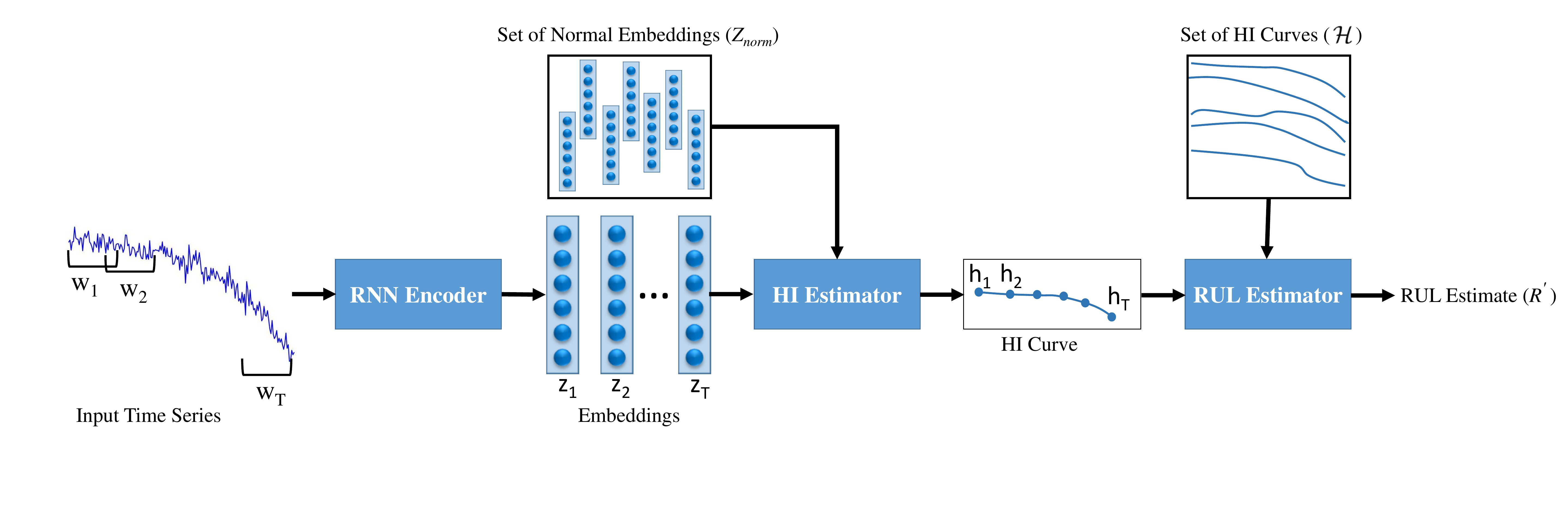}}
\caption{\label{fig:flow_chart}An overview of inference using Embed-RUL. The input time series is divided into windows. Each window is passed through an RNN Encoder to obtain its embedding. The embedding $z_t$ at time $t$ is compared with the embeddings in set Z$_{norm}$ of normal embeddings to obtain health estimate $h_t$ ($t=1,\dots,T$). The HI curve is then compared with HI curves of failed train instances in set $\mathcal{H}$ to obtain the RUL estimate $R'$.}
\end{figure*}

\subsubsection{Multilayered RNN with Dropout}\label{sec:gru_missing}
We use Gated Recurrent Units \citep{cho2014learning} in the hidden layers of sequence-to-sequence autoencoder. Dropout is used for regularization \citep{pham2014dropout,srivastava2014dropout} and is applied only to the non-recurrent connections, ensuring information flow across time-steps. 
For a multilayered RNN with $L$ hidden layers, the hidden state $\mathbf{z}_{t}^{l}$ at time $t$ for $l^{th}$ hidden layer is obtained from $\mathbf{z}_{t-1}^{l}$ and $\mathbf{z}_{t}^{l-1}$ as in Equation~\ref{eq:reset_gate}. 
The time series goes through the following transformations iteratively for $t=1$ through $T$, where $T$ is length of the time series:
\begin{equation}\label{eq:reset_gate}
\begin{aligned}
 reset\, gate: \mathbf{r}_t^l = \sigma(\mathbf{W}_r^l\cdot[\mathbf{D}(\mathbf{z}_{t}^{l-1}),\mathbf{z}_{t-1}^l])\\
 update\, gate: \mathbf{u}_t^l = \sigma(\mathbf{W}_u^l\cdot[\mathbf{D}(\mathbf{z}_{t}^{l-1}),\mathbf{z}_{t-1}^l])\\
 proposed\, state: \mathbf{\tilde z}_t^l = \tanh(\mathbf{W}_p^l\cdot[\mathbf{D}(\mathbf{z}_{t}^{l-1}),\mathbf{r}_t\odot \mathbf{z}_{t-1}^l])\\
 hidden\, state: \mathbf{z}_t^l = (1-\mathbf{u}_t^l)\odot \mathbf{z}_{t-1}^l+\mathbf{u}_t^l\odot\mathbf{\tilde z}_t^l\\
\end{aligned}
\end{equation}
where $\odot$ is Hadamard product, $[\mathbf{a},\mathbf{b}]$ is concatenation of vectors $\mathbf{a}$ and $\mathbf{b}$, $\mathbf{D}(\cdot)$ is dropout operator that randomly sets the dimensions of its argument to zero with probability equal to dropout rate, $\mathbf{z}_{t}^{0}$ equals the input at time $t$. $\mathbf{W}_r$, $\mathbf{W}_u$, and $\mathbf{W}_p$ are weight matrices of appropriate dimensions s.t. $\mathbf{r}_t^l, \mathbf{u}_t^l, \mathbf{\tilde z}_t^l$, and $\mathbf{z}_t^l$ are vectors in $\mathbb{R}^{c^l}$, where $c^l$ is the number of units in layer $l$. 
The sigmoid ($\sigma$) and $tanh$ activation functions are applied element-wise. 

\section{RUL Estimation using Embeddings}\label{sec:Embed-RUL}
We consider a scenario where sensor readings over the operational life of one or multiple instances of a machine or a component are available. We denote the set of instances by $\mathcal{I}$.
For an instance $i\in \mathcal{I}$, we consider a multi-sensor time series $\mathbf{x}^{(i)} =\{\mathbf{x}_1^{(i)},\mathbf{x}_2^{(i)},\cdots,\mathbf{x}_{T^{(i)}}^{(i)}\},$ where $T^{(i)}$ is the length of the time series, $\mathbf{x}_t^{(i)} \in \mathbb{R}^n$ is an $n$-dimensional vector corresponding to the readings of the $n$ sensors at time $t$. 
For a failed instance $i$, the length $T^{(i)}$ corresponds to the total operational life (from start to end of life) while for a currently operating instance the length $T^{(i)}$ corresponds to the elapsed operational life till the latest available sensor reading.

Typically, if $T^{(i)}$ is large, we divide the time series into windows (subsequences) of fixed length $w$. 
We denote a time series window from time $t_1$ to $t_2$ for instance $i$ by $\mathbf{x}^{(i)}(t_1,t_2)$.
A fixed-dimensional representation or embedding for each such window is obtained using an RNN Encoder that is trained in an unsupervised manner using RNN-ED. 
We train RNN-ED using time series subsequences from the entire operational life of machines (including normal as well as faulty operations)\footnote{Unlike the proposed approach, Recon-RUL \citep{malhotra2016multi} uses time series subsequences only from normal operation of the machine.}. 
We use the embedding for a window to estimate the health of the instance at the end of that window.
The RNN Encoder is likely to retain the important characterstics of machine behavior in the embeddings, and therefore discriminate between embeddings of windows corresponding to degraded behavior from those of normal behavior. 
We describe how these embeddings are obtained in Section~\ref{sec:RNN-ED_Missing}, and then describe how health index curves and RUL estimates can be obtained using the embeddings in Sections~\ref{sec:HI-curves} and~\ref{sec:RUL_estimation}, respectively. Figure~\ref{fig:flow_chart} provides an overview of the steps involved in the proposed approach for RUL estimation. 

\subsection{Obtaining Embeddings using RNN Encoder-Decoder}\label{sec:RNN-ED_Missing}
\begin{figure}
{\includegraphics[trim={0.8cm 10cm 0cm 4cm},scale=0.3,clip]{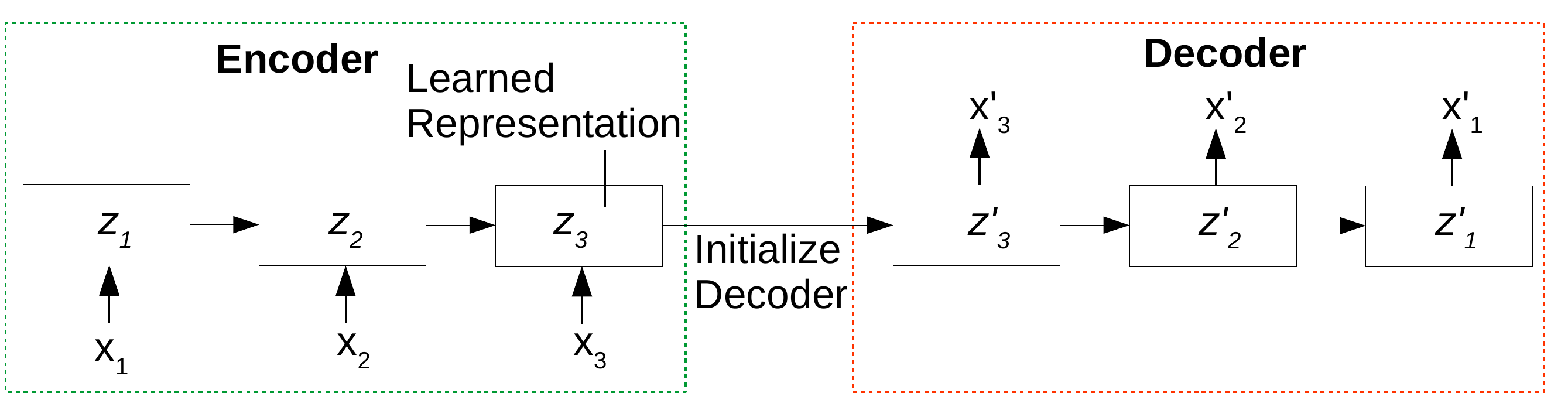}}
\caption{\label{fig:RNN-ED} RNN Encoder Decoder for toy time series $\{\mathbf{x}_1,\mathbf{x}_2,\mathbf{x}_3\}$.}
\end{figure}

We briefly introduce RNN Encoder-Decoder (RNN-ED) networks based on sequence-to-sequence (seq2seq) learning framework.
In general, a seq2seq model consists of a pair of multilayered RNNs trained together: an encoder RNN and a decoder RNN.
Figure~\ref{fig:RNN-ED} shows the workings of encoder-decoder pair for a sample time series $\{\mathbf{x}_1,\mathbf{x}_2,\mathbf{x}_3\}$.
Given an input time series $\mathbf{x}^{(i)}(t-w+1,t)$, the encoder RNN iterates through the points in the time series to compute the final hidden state $\mathbf{z}_{t}^{(i)}$, given by the concatenation of the hidden state vectors from all the layers in the encoder, s.t. $\mathbf{z}_{t}^{(i)}=[\mathbf{z}_{t,1}^{(i)},\mathbf{z}_{t,2}^{(i)},\dots,\mathbf{z}_{t,L}^{(i)}]$, where $\mathbf{z}_{t,l}^{(i)}$ is the hidden state vector for the $l^{th}$ layer of encoder. The total number of recurrent units in the encoder is given by $c =\sum_{l=1}^{L}c^l$, s.t. $\mathbf{z}_{t}^{(i)} \in \mathbb{R}^c$ (refer Section~\ref{sec:gru_missing}).

The decoder RNN has the same network structure as the encoder, and uses the hidden state $\mathbf{z}_{t}^{(i)}$ as its initial hidden state, and iteratively (for $w$ steps) goes through the transformations in Equation~\ref{eq:reset_gate} (followed by a linear output layer) to reconstruct the input time series. The overall process can be thought of as a non-linear mapping of the input multivariate time series to a fixed-dimensional vector representation (embedding) via an encoder function $f_{enc}$, followed by another non-linear mapping of the fixed-dimensional vector to a multivariate time series via a decoder function $f_{dec}$:
\begin{equation}
\begin{aligned}
embedding \quad \mathbf{z}_t^{(i)}=f_{enc}(\mathbf{x}^{(i)}(t-w+1,t))\\
reconstructed\:\,time\:\,series \quad \mathbf{x}'^{(i)}(t-w+1,t)=f_{dec}(\mathbf{z}_t^{(i)})\\
\end{aligned}
\end{equation}
The reconstruction error at any point $t'$ in $(t-w+1),  \dots, t$ is $\quad \mathbf{e}_{t'}^{(i)}=\mathbf{x}^{(i)}_{t'}-\mathbf{x}'^{(i)}_{t'}$. The overall reconstruction error for the input time series window $\mathbf{x}^{(i)}(t-w+1,t)$ is given by $e_t^{(i)}=\sum_{t'=(t-w+1)}^{t}\parallel \mathbf{e}_{t'}^{(i)} \parallel_2^{2}$. The RNN-ED is trained to minimize the loss function given by the squared reconstruction error: $\mathcal{L}=\sum_{i\in\mathcal{I}}\sum_{t=w}^{T^{(i)}} {e_t^{(i)}}$.

\begin{figure}[h]
\includegraphics[trim={2cm 0.0cm 2.0cm 1.5cm},scale=0.23,clip]{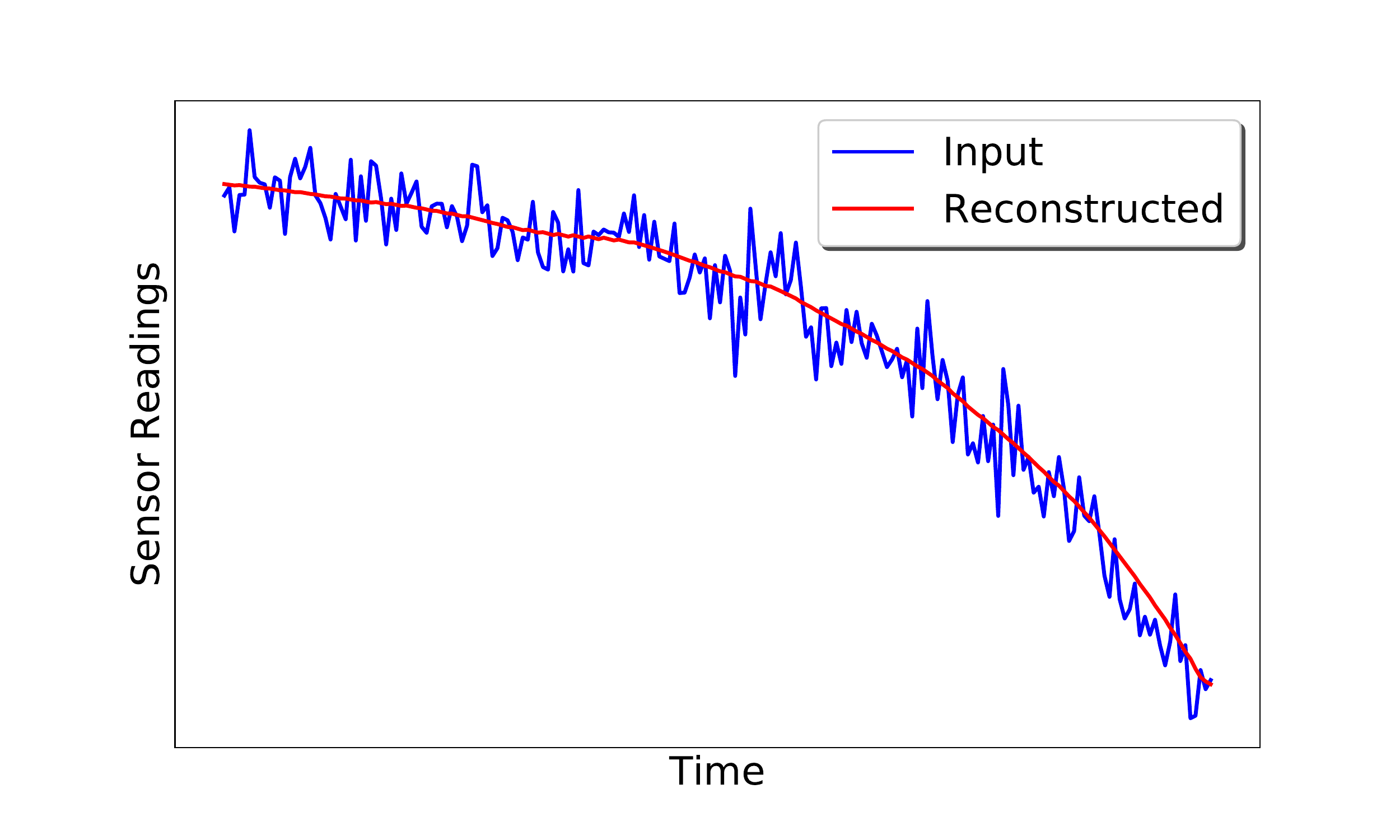}
\caption{\label{fig:RNN-ED-In-Out}Motivating example showing input time series and its reconstruction using RNN-ED. Sample taken from results on Turbofan Engine dataset (Section~\ref{sec:performance_Turbofan}).}
\end{figure}

Typically, along with the final hidden state, an additional input is given to the decoder RNN at each time step. This input is the output of the decoder RNN at the previous time step, as used in \cite{malhotra2016multi}. We, however, do not give any such additional inputs to the decoder along with the final hidden state of encoder. This ensures that the final hidden state of encoder retains all the information required to reconstruct the time series back via the decoder RNN. 
This approach of learning robust embeddings or representations for time series has been shown to be useful for time series classification in \cite{malhotra2017timenet}. 
Figure~\ref{fig:RNN-ED-In-Out} shows a typical example of input and output from RNN-ED, where the smoothed reconstruction suggests that the embeddings capture the necessary pattern in the input and remove noise. 

\subsubsection{Handling missing values}\label{sec:handling-missing}
In real-world data, the sensor readings tend to be intermittently missing. We include  \textit{masking} and \textit{delta} vectors as additional inputs to the RNN-ED at each time instant, (as in \cite{che2016recurrent}). The masking vector helps to identify the sensors that are missing at time $t$, and the delta vector indicates the time elapsed till $t$, from the most recent non-missing values for sensors in the past. 
We omit superscript $(i)$ for denoting an instance of the machine from the notation of masking and delta vectors defined below for simplicity.
\begin{itemize}
\item \textit{Masking vector} ($\mathbf{m}_t$) denotes the missing sensors at time $t$ and $\mathbf{m}_t \in \{0,1\}^{n}$, where $n$ is the number of sensors. The $j^{th}$ element of vector $\mathbf{m}_t$ is given by:
\begin{equation}
       m_t^j = 
        \begin{cases}
            0, & \text{if $x_t^j$ is missing} \\
            1, & \text{otherwise}
        \end{cases}
    \end{equation}
    where  $j=1,\ldots,n$, and $x_t^j$ denotes the $j^{th}$ element of vector $\mathbf{x}_t$.
When $m_t^j=0$, we set $x_t^j$ to 0 or to the average value for $j^{th}$ sensor (we use 0 for the experiments in Section~\ref{sec:results}).
\item \textit{Delta vector} ($\boldsymbol{\delta}_t$) indicates the time elapsed till $t$, from the most recent non-missing values for the sensors in the past and $\boldsymbol{\delta}_t \in \mathbb{R}^n$. 
The $j^{th}$ element of vector $\boldsymbol{\delta}_t$ is given by: 
\begin{equation}
      \delta_t^j = 
        \begin{cases}
            y_t-y_{t-1}+\delta_{t-1}^j, & \text{if $t>1,m_{t-1}^j=0$ } \\
            y_t-y_{t-1}, & \text{if $t>1,m_{t-1}^j=1$ } \\
            0, & \text{for $t=1$}
        \end{cases}
    \end{equation}
where  $j=1,\ldots,n$ and $y_t \in \mathbb{R}$ is the time elapsed from start when $t^{th}$ reading is available and $y_1=0$. 
It is to be noted that the sensor readings may not be available at regular time intervals. Therefore, 
the sequence of readings is indexed by time $t=1,2,\dots,T$, while the actual timestamps are denoted by $y_1,y_2,\dots,y_T$.
\end{itemize}

The masking and delta vectors are given as additional inputs to the RNN-ED but are not reconstructed, s.t. only the actual sensors are reconstructed. 
Therefore, the modified input time series
$\mathbf{\hat{x}}_{t}^{(i)}=[\mathbf{x}_{t}^{(i)},\mathbf{m}_{t}^{(i)},\boldsymbol{\delta}_t^{(i)}]$, while the corresponding target to be reconstructed is $\mathbf{x}_{t}^{(i)}$.
The loss function $(\mathcal{L})$ of the RNN-ED is also modified accordingly, so that the model is not penalized for reconstructing the missing sensors incorrectly. 
The contribution of a time series subsequence $\mathbf{x}^{(i)}(t-w+1,t)$ to the loss function is thus given by $e_t^{(i)}=\sum_{t'=(t-w+1)}^{t}\parallel \mathbf{e}_{t'}^{(i)}\cdot \mathbf{m}_{t'}^{(i)}\parallel_2$. 
In effect, the network focuses on reconstructing the available sensor readings only.

\subsection{Obtaining HI Curves using Embeddings}\label{sec:HI-curves}
Here we describe how the embeddings of time series subsequences are utilized to estimate the health of machines. 
Since the RNN Encoder captures the important patterns in the input time series subsequences, the embeddings thus obtained can be used to differentiate between normal and degraded regions in the data. 
We maintain a set of embeddings, $Z_{norm}$, corresponding to the time series subsequences from the normal behavior of all the instances in $\mathcal{I}$. As a machine operates, its health degrades over time and the corresponding subsequence embeddings tend to be different from those in $Z_{norm}$. So, we estimate the HI for a subsequence as follows:
\begin{equation}\label{eq:Health_index}
h_t^{(i)}=min(\parallel \mathbf{z}_t^{(i)}-\mathbf{z}\parallel_{2}), \;\forall  \:\mathbf{z} \:\in Z_{norm}
\end{equation}
 
The HI curve for an instance $i$ obtained from the HI estimates at each time is denoted by $h^{(i)}=\{h^{(i)}_w,h^{(i)}_{w+1},\dots,h^{(i)}_{T^{(i)}}\}$. 
Like the set of normal embeddings $Z_{norm}$, we also maintain a set $\mathcal{H}$ containing the HI curves of all instances in $\mathcal{I}$.

It is to be noted that HI values are usually assumed to have value between 0 and 1, where 0 means very poor health and 1 means perfect normal health (as shown in Figure \ref{fig:Curve_matching}). The HI as defined in Equation \ref{eq:Health_index} follows inverse definition, i.e. it is low for normal health and high for poor health (as shown in Figure \ref{fig:toy_example}). This can be easily transformed to adhere to the standard range of 0-1 through suitable normalization/scaling procedure if required, as used in \cite{malhotra2016multi}.

\subsection{RUL Estimation using HI Curves}\label{sec:RUL_estimation}
We use the same approach for estimating RUL from the HI curve as in \cite{malhotra2016multi}.
We present it here for the sake of completeness.
To estimate the RUL for a test instance $i^*$, its HI curve $h^{(i^*)}$ is compared with the HI curves in $\mathcal{H}$.
The initial health of a train instance and a test instance need not be same. We therefore allow for a time-lag $t_D$ in comparing the HI curve of test instance and train instance.

The similarity between the HI curves of the test instance $i^*$ and a train instance $i\in \mathcal{I}$ for a time-lag $t_D$ is given by: \begin{equation}\label{eq:sim}
 s(i^*,i,t_D)=exp(-\frac{1}{T^{(i^*)}}\sum_{k=w}^{T^{(i^*)}}{(h_k^{(i^*)}-h_{k+t_D}^{(i)})^2}/\lambda)
\end{equation}
$\lambda>0$, 
$t_D \in \{1,2,...,\tau\}$, 
$t_D+T^{(i^*)} \leq T^{(i)}$. 
Here, $\tau$ is maximum allowed time-lag, and $\lambda$ controls the notion of similarity: a small value of $\lambda$ would imply a large value for $s$ even when the difference in HI curves is small. The RUL estimate for $i^*$ based on the HI curve for $i$ and time-lag $t_D$ is given by $R'^{(i^*)}(i,t_D)=T^{(i)}-T^{(i^*)}-t_D$.

A weighted average of the RUL estimates obtained using all combinations of $i$ and $t_D$ is used as the final estimate $R'^{(i^*)}$, and is given by:
\begin{equation}\label{eq:rul_estimate}
 R'^{(i^*)}=\frac{\sum{s(i^*,i,t_D)\cdot R'^{(i^*)}(i,t_D)}}{\sum{s(i^*,i,t_D)}}
\end{equation}
where the summation is over only those combinations of $i$ and $t_D$ which satisfy $s(i^*,i,t_D)\geq \alpha.s_{max}$, where $s_{max}=max_{i\in \mathcal{I},t_D \in \{1\:,\dots,\:\tau\}}$ $\{s(i^*,i,t_D)\}$, $0\leq \alpha \leq 1$. 


\section{Experimental Evaluation}\label{sec:results}
We evaluate our proposed approach for RUL estimation on two datasets: i) a publicly available C-MAPSS Turbofan Engine dataset \cite{saxena2008turbofan}, ii) a proprietary real-world pump dataset.
We use Tensorflow \cite{abadi2016tensorflow} library for implementing the various RNN models.
We present the details of datasets in Section~\ref{sec:data_desc}. 
In Section~\ref{sec:rul_eval}, we show that the results for embedding distance based approaches for RUL estimation compare favorably to the previously reported results using reconstruction error based approaches \cite{malhotra2016multi} on the engine dataset , as well as on the real-world pump dataset.
Further, we evaluate the robustness of the embedding distances and reconstruction error based approaches by measuring the effect of additive random Gaussian noise in the sensor readings on RUL estimation in Section~\ref{sec:noise_test}. 

\subsection{Datasets Description\label{sec:data_desc}}
\subsubsection{Engine dataset}
We use the first dataset from the four simulated turbofan engine datasets from the NASA Ames Prognostics Data Repository \cite{saxena2008turbofan}. 
This dataset contains time series of readings for 24 sensors for 100 train instances (\textit{train\_FD001.txt}) of turbofan engine from the beginning of usage till end of life. 
There are 100 test instances for which the time series are pruned some time prior to failure, s.t. the instances are currently operational and their RUL needs to be estimated (\textit{test\_FD001.txt}).
The actual RUL for the test instances are available in \textit{RUL\_FD001.txt}. 
Noticeably, each engine instance has a different initial degree of wear such that the initial HI of each instance is likely to be different (implying potential usefulness of $\tau$ as introduced in Section~\ref{sec:RUL_estimation}).

We randomly select 80 train instances to train the models. 
Remaining 20 instances from the train set are used as validation set to select the parameters. 
The trajectories for these 20 engines are randomly truncated at five different locations to obtain five different instances from each instance for the RUL estimation task.
We use Principal Components Analysis (PCA) \cite{jolliffe2002principal} to reduce the dimensionality of data and select the number of principal components ($p$) to be used based on the validation set.

\subsubsection{Pump dataset}
This dataset contains hourly sensor readings for 38 pumps that have reached end of life and 24 pumps that are currently operational. This dataset contains readings over a period of 2.5 years with each pump having 7 sensors installed on it.
The 38 failed instances are randomly split into training, validation and test sets with 70\%, 15\%, and 15\% instances in them, respectively. 
The 24 operational instances are added to training and validation set only for obtaining the RNN-ED model (they are not part of the set $\mathcal{H}$ as their actual RUL is not known). 
The data is notably sparse with over 45\% missing values across sensors.
Also, for most pumps the sensor readings are not available from the date of installation but only few months (average 3.5 months) after the installation date. 
Depending on the time elapsed, the health degradation level when sensor data is first available for each pump varies significantly.
The total operational life of the pumps varies from a minimum of 57 days to a maximum of 726 days.

We downsample the time series data from the original one reading per hour to one reading per day. To do this, we use following four statistics for each sensor over a day as derived sensors: minimum, maximum, average, and standard deviation, such that there are 28 (=$7\times 4$) derived sensors for each day. Further, using the derived sensors also helps take care of missing values which reduce from 45\% for hourly sampling rate data to 33\% for daily sampling rate data.
We use masking and delta vectors as additional inputs in this case to train RNN-ED models as described in Section~\ref{sec:handling-missing}, s.t. the final input dimension is 42 (28 for derived sensors, and 7 each for masking and delta vectors).
Unlike the engine dataset where RUL is estimated only at the last available reading for each test instance, here we estimate RUL on every third day of operation for each test instance.

A description of the performance metrics used for evaluation (taken from \cite{saxena2008metrics}) is provided in Appendix~\ref{sec:metrics}. The hyper-parameters of our model, to be tuned are: number of principal components ($p$), number of hidden layers for RNN-ED ($L$), number of units in a hidden layer $l$ ($c^l$) (we use same number of units in each hidden layer), dropout rate ($d$), window length ($w$), maximum allowed time-lag ($\tau$), similarity threshold ($\alpha$), maximum predicted RUL ($R_{max}$), and parameter ($\lambda$). The window length ($w$) can be tuned as a hyperparameter but in practice domain knowledge based selection of window length may be effective.

\subsection{Embeddings for RUL Estimation\label{sec:rul_eval}}
We follow similar evaluation protocol as used in \cite{malhotra2016multi}.
To the best of our knowledge, the reconstruction error based model, LR-ED$_2$, reported the best performance for RUL estimation on the engine dataset in terms of timeliness score (refer Appendix~\ref{sec:benchmarks}).
We compare variants of embedding distance based approach and reconstruction error based approach. 
We refer to HI curve obtained using the proposed embedding distance based approach as HI$_e$ (refer Section~\ref{sec:HI-curves}), and the HI curve obtained using the reconstrcution error based approach in \cite{malhotra2016multi} as HI$_r$.
Here, we refer the reconstruction error based LSTM-ED, LR-ED$_1$ and LR-ED$_2$ models reported in \cite{malhotra2016multi}, as Recon-RUL, Recon-LR$_1$, and Recon-LR$_2$, respectively. 
We compare following models based on RNNs for RUL estimation task: 
\begin{itemize}
\item \textit{Embed-RUL Vs Recon-RUL}: We compare RUL estimation performance of Embed-RUL that uses HI$_e$ curves and Recon-RUL that uses HI$_r$ curves.
\item \textit{Linear Regression models}: We learn a linear regression model (as described in Section~\ref{sec:lin_reg_hi}) using normalized health index curves HI$_e$ as target and call it as Embed-LR$_1$. Embed-LR$_2$ is obtained using squared normalized HI$_e$ as target for the linear regression model. Similarly, Recon-LR$_1$ and Recon-LR$_2$ are obtained based on HI$_r$.

\item \textit{RNN Regression model}: RNN-based regression model (RNN-Reg.) is directly used to predict RUL (similar to \cite{heimes2008recurrent})
\end{itemize}
\begin{figure}[b]
\subfigure[Engine dataset\label{fig:turbofan_tSNE}]{\includegraphics[trim={3cm 2cm 3cm 2.5cm},scale=0.25,clip]{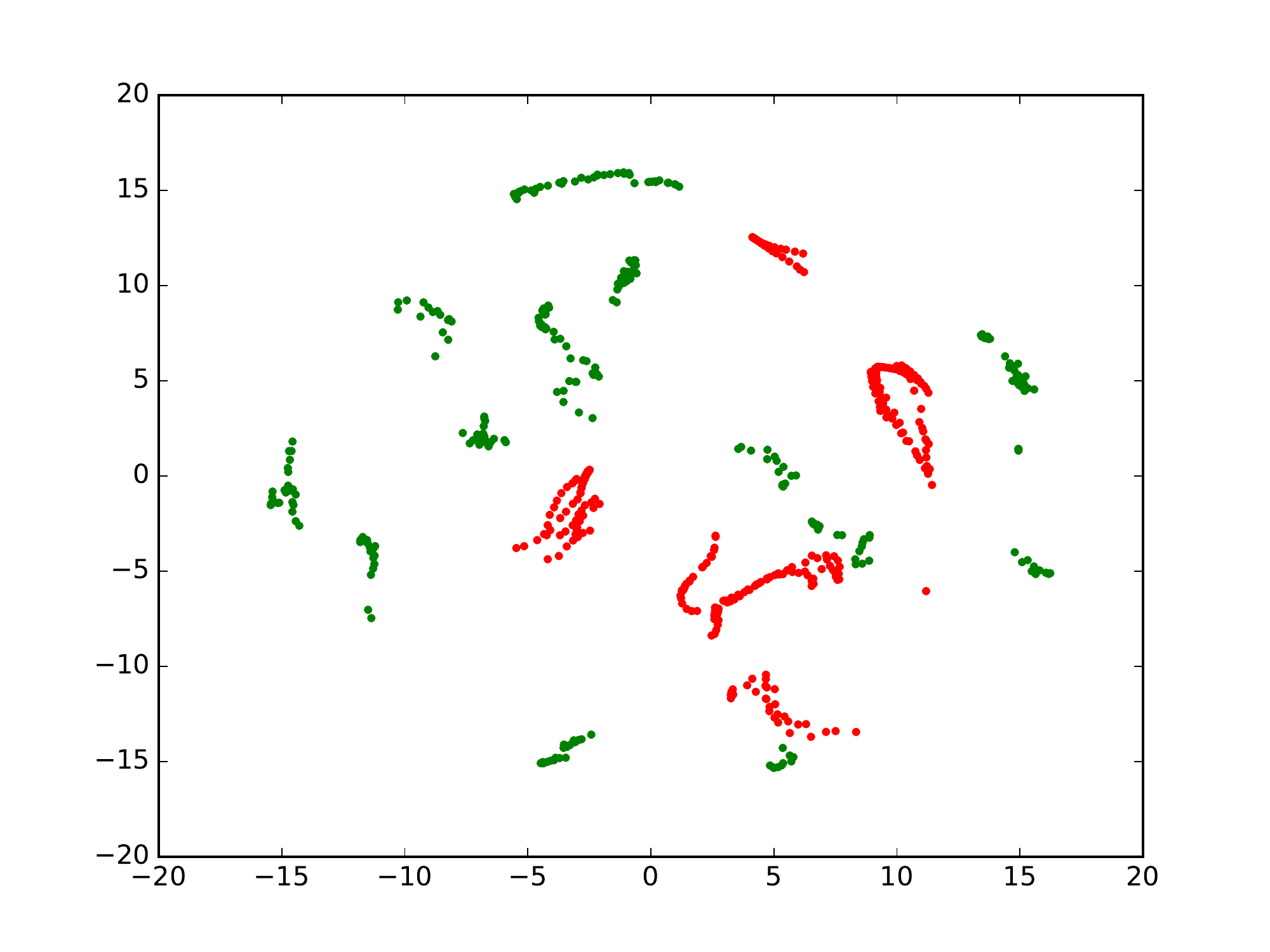}}
\hspace{25pt}
\subfigure[Pump dataset \label{fig:pump_tSNE}]{\includegraphics[trim={3cm 2cm 3cm 2.5cm},scale=0.25,clip]{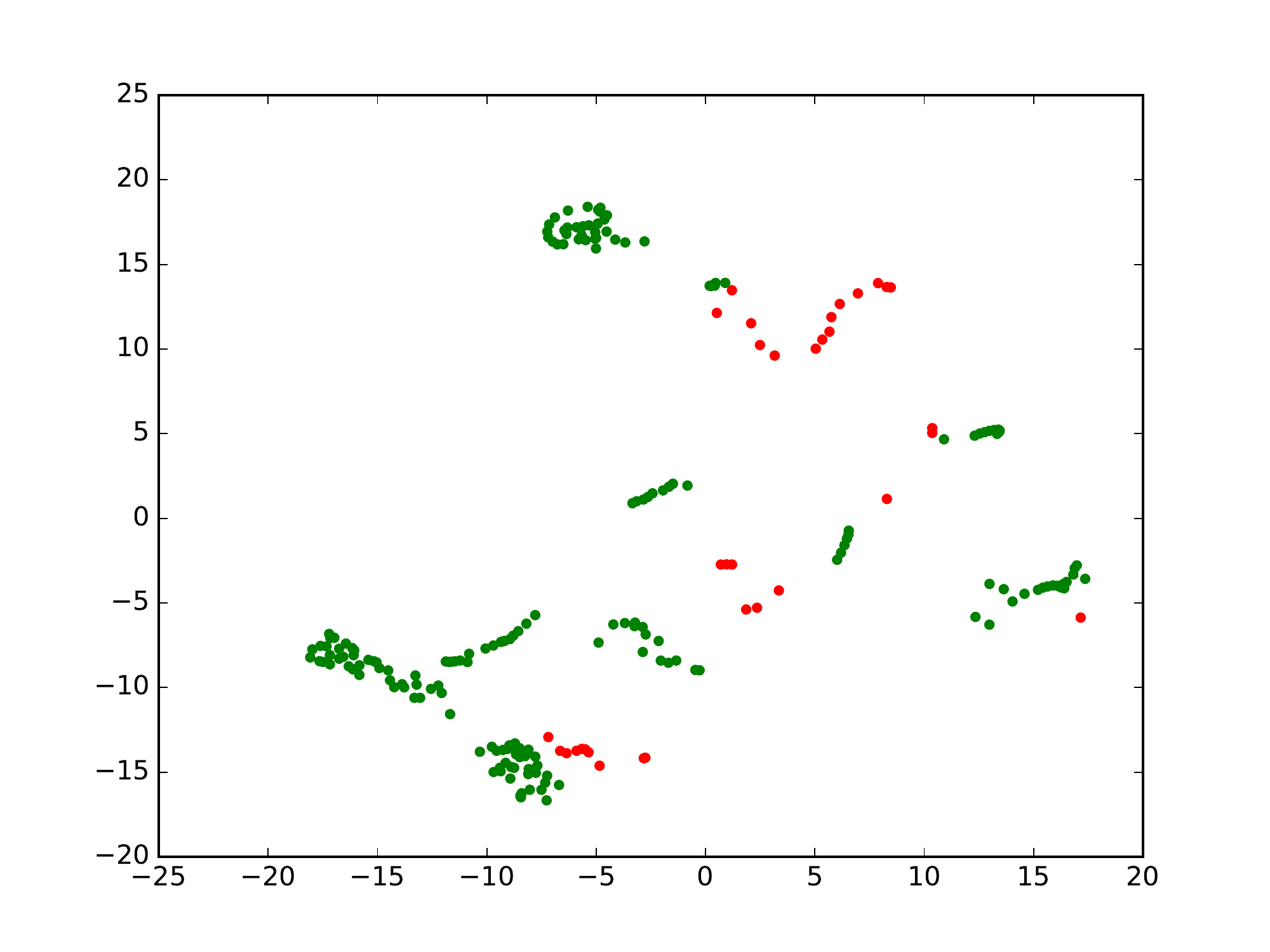}}
\caption{\label{fig:tSNE_plots}t-SNE plot for embeddings of normal (green) and degraded (red) windows}
\end{figure}
\begin{table}[bp]
\footnotesize
\begin{tabularx}{0.48\textwidth}{@{}ccccc@{}}
\toprule
&\multicolumn{2}{c}{\bfseries Engine Dataset}
&\multicolumn{2}{c}{\bfseries Pump Dataset} \\
\cmidrule(l){2-3} \cmidrule(l){4-5} 
Noise ($\sigma$)&Recon-RUL 
&Embed-RUL
&Recon-RUL 
&Embed-RUL\\
&&(proposed)&&(proposed)\\
\midrule
0.0 &546&456&1979&1304 \\
0.1&548&462&2003& 1298 \\
0.2&521&478&2040&1293\\
0.3&523&460&2068&1259 \\
0.4&484&473&2087&1280\\
\hline
Mean&524&\textbf{466}&2035&\textbf{1287}\\
Std. Dev.&23&\textbf{8}&40&\textbf{16}\\
\bottomrule
\end{tabularx}
\caption{Robustness Evaluation: MSE values \label{tab:noise_results}}
\end{table}

\subsubsection{Performance on Engine dataset}\label{sec:performance_Turbofan}
\begin{table*}[t]
\footnotesize
\begin{tabular}{c|cc|cc|cc|c}
\toprule
Metric&Recon-RUL&Embed-RUL&Recon-LR$_1$&Embed-LR$_1$&Recon-LR$_2$&Embed-LR$_2$&RNN-Reg.\\
&&(proposed)&&(proposed)&&(proposed)&\\
\midrule
S&1263&\textbf{810}&477&\cellcolor{lightgray}\textbf{219}&256&\textbf{232}&352\\
MSE&546&\textbf{456}&288&\textbf{155}&\textbf{164}&167&219\\
A(\%)&36&\textbf{48}&\textbf{65}&59&\textbf{67}&62&64\\
MAE&18&\textbf{17}&12&\textbf{10}&\textbf{10}&\textbf{10}&11\\
\tablefootnote{Referred to as MAPE$_1$ in \cite{malhotra2016multi}}
MAPE&\textbf{39}&\textbf{39}&20&\textbf{19}&\textbf{18}&19&17\\
FPR(\%)&34&\textbf{23}&19&\textbf{14}&\textbf{13}&15&22\\
FNR(\%)&30&\textbf{29}&\textbf{16}&27&\textbf{20}&23&24\\
\bottomrule
\end{tabular}
\caption{Engine dataset: Performance comparison \label{tab:turbofan_results}}
\end{table*}

\begin{table*}[t]
\footnotesize
\begin{tabular}{c|cc|cc|cc|c}
\toprule
Metric&Recon-RUL&Embed-RUL&Recon-LR$_1$&Embed-LR$_1$&Recon-LR$_2$&Embed-LR$_2$&RNN-Reg.\\
&&(proposed)&&(proposed)&&(proposed)&\\
\midrule
MSE&1979& \cellcolor{lightgray}\textbf{1304}&\textbf{2277}&2288&2365&\textbf{2312}&3422\\
MAE&40&\textbf{33}&\textbf{38}&42&\textbf{38}&42&48\\
\bottomrule
\end{tabular}
\caption{Pump dataset: Performance comparison \label{tab:pump_results} (We only consider MSE and MAE metrics for this dataset as there are no standard $\tau_1$ and $\tau_2$ known for this dataset to compute the other metrics.)}
\end{table*}
\begin{figure*}[thp]
{\includegraphics[scale=0.2]{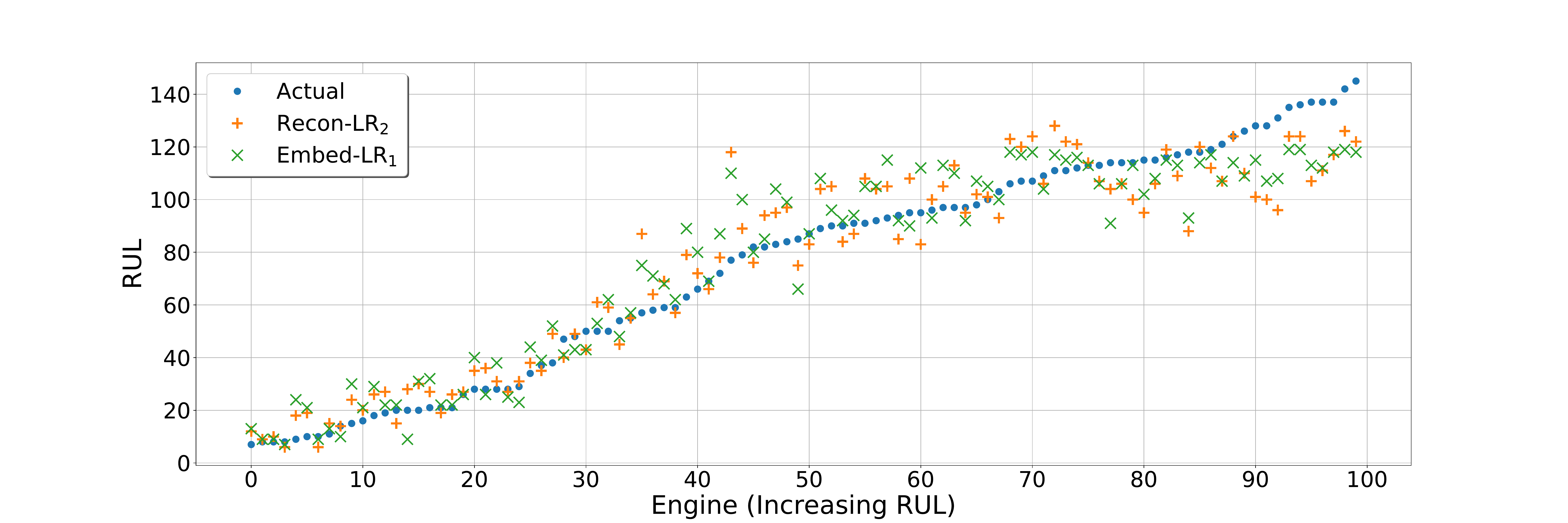}}
\caption{\label{fig:turbofan_predictions}Engine dataset: RUL estimates given by Embed-LR$_1$ and Recon-LR$_2$.}
\end{figure*}

\begin{figure*}[th]
\subfigure[\label{fig:Min_MSE}Pump for which Embed-RUL MSE is minimum.]{\includegraphics[width=\columnwidth]{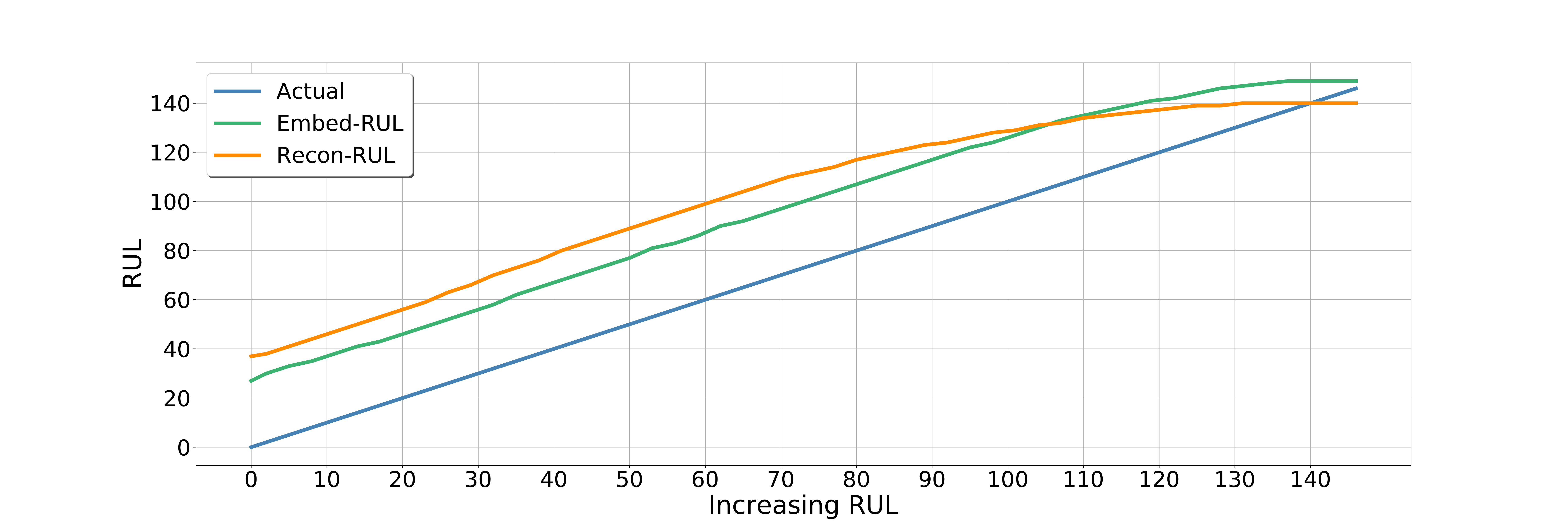}}
\subfigure[\label{fig:Max_MSE}Pump for which Embed-RUL MSE is maximum.]
{\includegraphics[width=\columnwidth]{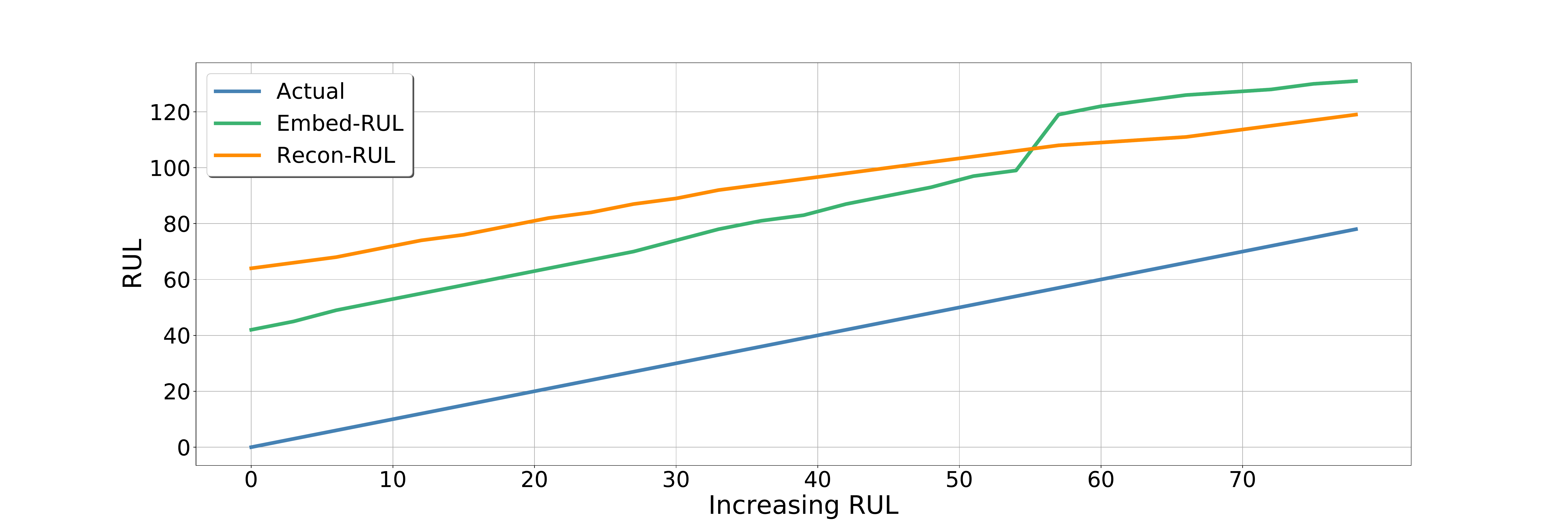}}
\caption{\label{fig:pump_predictions}{Pump dataset: RUL estimates given by Embed-RUL and Recon-RUL.}}
\end{figure*}
\begin{figure*}[t!]
\subfigure[Embed-RUL Vs Recon-RUL.\label{fig:raw_hist_turbofan}]{\includegraphics[scale=0.22]{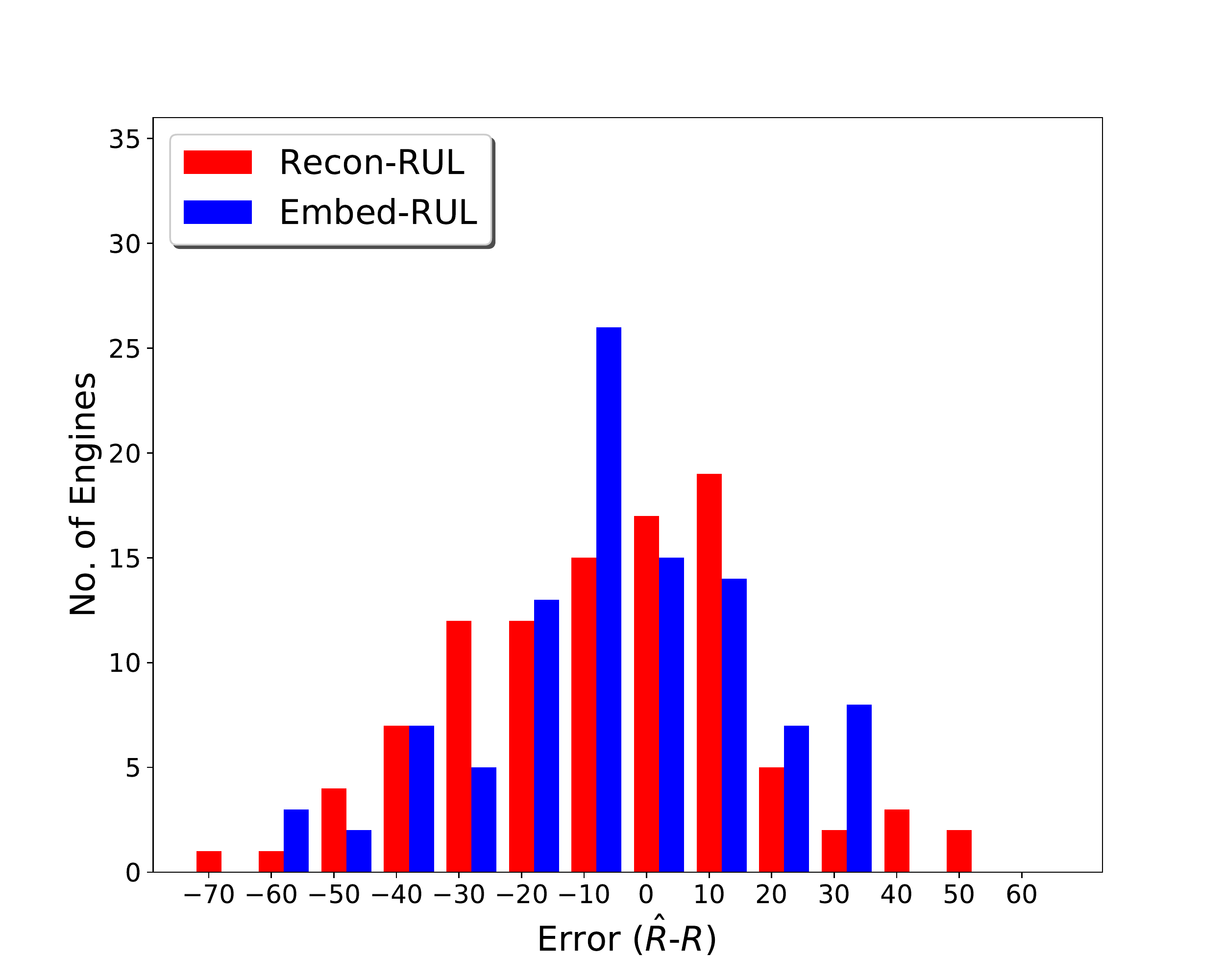}}
\subfigure[Embed-LR$_1$ Vs Recon-LR$_2$. \label{fig:LR_hist_turbofan}]{\includegraphics[scale=0.22]{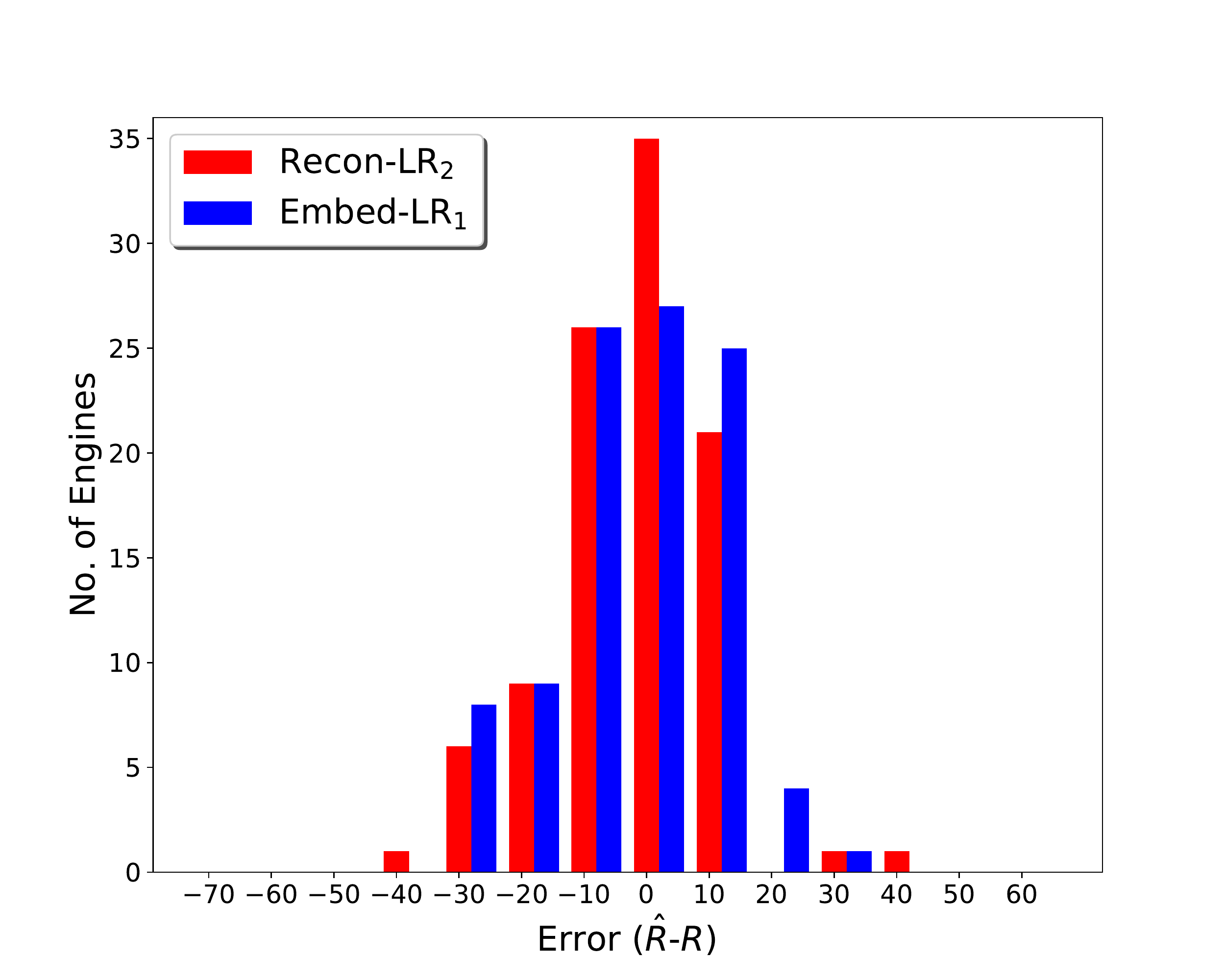}}
\caption{\label{fig:error_hist_turbofan}Engine dataset: Histograms of prediction errors}
\end{figure*}

We use $\tau_1$=13, $\tau_2$=10 as proposed in \cite{saxena2008damage} for this dataset (refer Equations~\ref{eq:timeliness}-\ref{eq:mape-1} in Appendix~\ref{sec:metrics}). The parameters are obtained using grid search to minimize the timeliness score $S$ (refer Equation~\ref{eq:timeliness}) on the validation set. 
The parameters obtained for the best model (Embed-LR$_{1}$) are $p=2$, $L=1$, $c^l=55$, $d=0.2$, $w=30$, $\tau=30$, $\alpha=0.95$, $R_{max}= 120$, and $\lambda = 0.005$.

Table~\ref{tab:turbofan_results} shows the performance in terms of various metrics on this dataset.
\textit{We observe that each variant of embedding distance based approach perform better than the corresponding variant of reconstruction error based approach in terms of timeliness score $S$.}
Figure~\ref{fig:raw_hist_turbofan} shows the distribution of errors for Embed-RUL and Recon-RUL models, and Figure~\ref{fig:LR_hist_turbofan} shows the distribution of errors for the best linear regression models of embedding distance (Embed-LR$_1$) and reconstruction error (Recon-LR$_2$) based approaches.
The error ranges for reconstruction error based models are more spread-out (e.g., -70 to +50 for Recon-RUL) than the corresponding embedding distances based models (e.g., -60 to +30 for Embed-RUL), suggesting the robustness of the embedding distances based models. Figure~\ref{fig:turbofan_predictions} shows the actual RULs and the RUL estimates from Embed-LR$_1$ and Recon-LR$_2$. 
\subsubsection{Performance on Pump dataset}\label{sec:pump_results}
The parameters are obtained using grid search to minimize the MSE for RUL estimation on the validation set.
The parameters for the best model (Embed-RUL) are $L=1$, $c^l=390$, $d=0.3$, $w=30$, $\tau=70$,  $\alpha=0.8$, $R_{max}= 150$, and $\lambda = 10$. 
The MSE and MAE performance metrics for the RUL estimation task are given in Table~\ref{tab:pump_results}.
\textit{The embedding distance based Embed-RUL model performs significantly better than any of the other approaches.}
It is $\approx 35\%$ better than the second best model (Recon-RUL).
The linear regression (LR) based approaches perform significantly worse than the raw embedding distance or reconstruction error based approaches for HI estimation indicating that the temporal aspect of the sensor readings is very important in this case.
Figure~\ref{fig:pump_predictions} shows the actual and estimated RUL values for the pumps with best and worst performance in terms of MSE for the Embed-RUL model. 


\subsubsection{Qualitative Analysis of Embeddings}\label{sec:embeddings_qualitative}
We analyze the embeddings given by RNN Encoder for the Embed-RUL models.
The original dimension of embeddings for Embed-RUL for engine and pump datasets are 55 and 390, respectively. 
We use t-SNE \cite{maaten2008visualizing} to map the embeddings to 2-D space.
Figure~\ref{fig:tSNE_plots} shows the 2-D scatter plot for the embeddings at the first 25\% (normal behavior) and last 25\% (degraded behavior) points in the life of all test instances. 
\textit{We observe that RNN Encoder tends to give different embeddings for windows corresponding to normal and degraded behaviors.} 
The scatter plots indicate that normal windows are close to each other and far from degraded windows, and vice-versa.
Note: Since t-SNE does non-linear dimensionality reduction, the actual distances between normal and degraded windows may not be reflected in these plots. 


\subsection{Robustness of Embeddings to Noise}\label{sec:noise_test}
\begin{figure}
\includegraphics[trim={2cm 0.0cm 2cm 1.5cm},scale=0.25,clip]{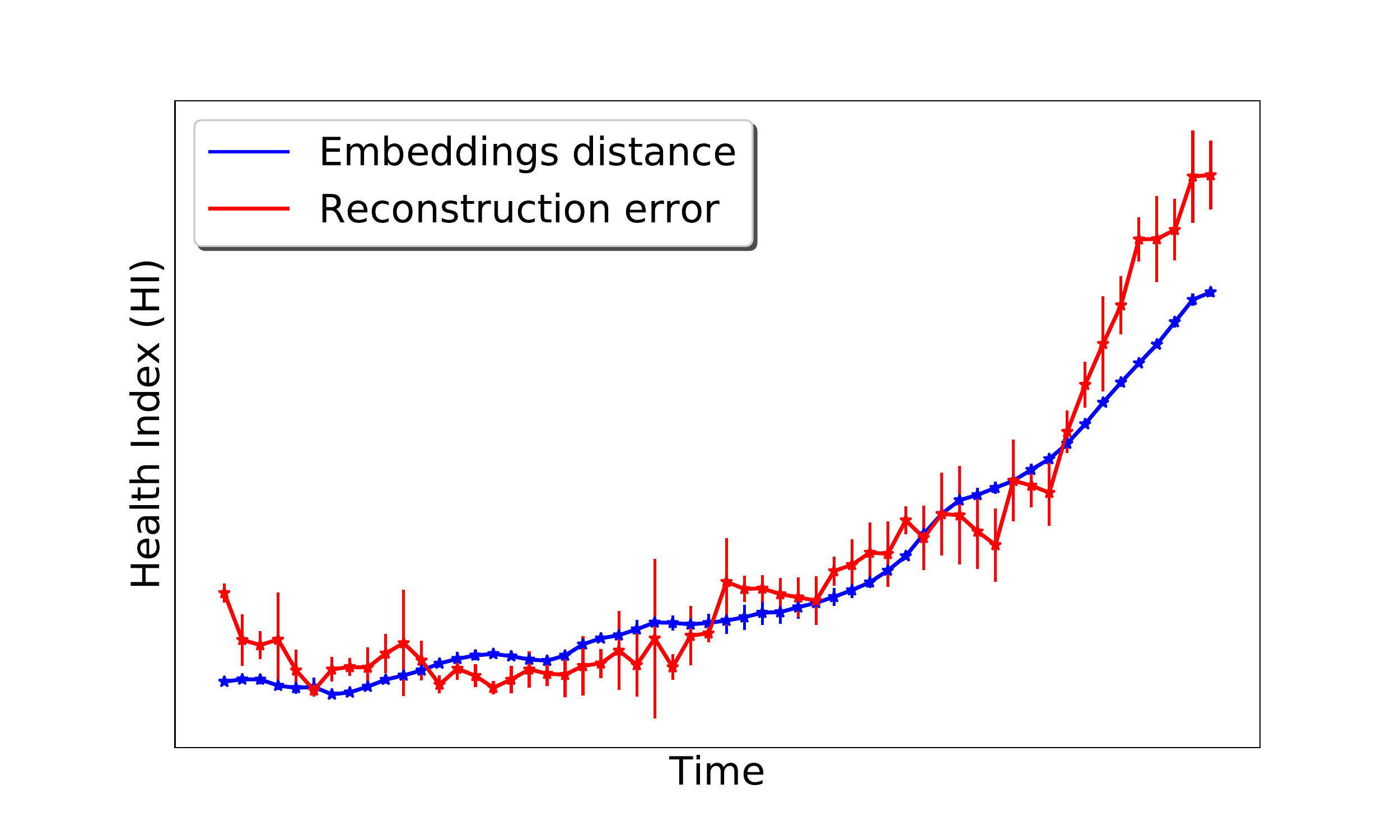}
\caption{\label{fig:toy_example} Impact of noise on Health Index (HI) for test engine \# 31}
\end{figure}
We evaluate the robustness of Embed-RUL and Recon-RUL for RUL estimation by adding Gaussian noise to the sensor readings. 
The sensor reading $\mathbf{x}_t^{(i^*)}$ for a test instance $i^*$ at any time $t$ is corrupted with additive Gaussian noise to obtain a noisy version $\mathbf{x'}_t^{(i^*)}$ s.t.
$\mathbf{x'}_t^{(i^*)}|\mathbf{x}_t^{(i^*)} \,\sim\,\mathcal{N}(\mathbf{x}_t^{(i^*)},\sigma^2I)$.

Table~\ref{tab:noise_results} shows the effect of noise on performance for both engine and pump datasets. 
\textit{For both datasets, the standard deviation of the MSE values over different noise levels is much lesser for Embed-RUL compared to Recon-RUL. This suggests that embedding distance based models are more robust to noise compared to reconstruction error based models.}
Also, for the engine dataset, we observe similar behavior in terms of timeliness score ($S$): $819\pm41$ for Embed-RUL and $1189\pm110$ for Recon-RUL.

Figure~\ref{fig:toy_example} depicts a sample scenario showing the health index generated from noisy sensor data. The vertical bar corresponds to 1-sigma deviation in estimate. The reconstruction error and embedding distance increase over time indicating gradual degradation. While reconstruction error based HI varies significantly with varying noise levels, embedding distance based HI is fairly robust to noise. This suggests that reconstruction error varies significantly with change in noise levels impacting HI estimates while distance between embeddings does not change much leading to robust HI estimates.

\section{Discussion}\label{sec:discussion}
We have proposed an approach for health monitoring via health index estimation and remaining useful life (RUL) estimation. The proposed approach is capable of dealing with several of the practical challenges in data-driven RUL estimation including noisy sensor readings, missing data, and lack of prior knowledge about degradation trends. The RNN Encoder-Decoder (RNN-ED) is trained in an unsupervised manner to learn fixed-dimensional representations or embeddings to capture machine behavior. The health of a machine is then estimated by comparing the recent embedding with the existing set of embeddings corresponding to normal behavior.
We found that our approach using RNN-ED based embedding distances is better compared to the previously known best approach using RNN-ED based reconstruction error on the engine dataset. 
The proposed approach also gives better results on the real-world pump dataset.
We have also shown that embedding distances based RUL estimates are robust to noise. 

\bibliographystyle{ACM-Reference-Format}
\bibliography{BibTex/phm-kdd2016,BibTex/phm-kdd2017,BibTex/sensor_analytics}

\appendix
\section{Performance metrics}\label{sec:metrics}
There are several metrics proposed to evaluate the performance of prognostics models \cite{saxena2008metrics}. We measure the performance of our models in terms of Timeliness Score (S), Accuracy (A), Mean Absolute Error (MAE), Mean Absolute Percentage Error (MAPE), False Positive Rate (FPR) and False Negative Rate (FNR) as mentioned in Equations~\ref{eq:timeliness}-\ref{eq:mape-1}.

For a test instance $i^*$, the error $\Delta^{(i^*)} =\hat{R}^{(i^*)} - {R}^{(i^*)}$ between the estimated RUL ($\hat{R}^{(i^*)}$) and actual RUL ($R^{(i^*)}$). The timeliness score $S$ used to measure the performance of a model is given by:

\begin{equation}\label{eq:timeliness}
 S=\sum^N_{i^*=1} (exp({\gamma\cdot|\Delta^{(i^*)}|})-1)
\end{equation}
where $\gamma=1/\tau_1$ if $\Delta^{(i^*)}<0$, else $\gamma=1/\tau_2$, $N$ is total test instances. Usually, $\tau_1 > \tau_2$ such that late predictions are penalized more compared to early predictions. The lower the value of $S$, the better is the performance.

\begin{equation}\label{eq:accuracy}
 A=\frac{100}{N}\sum^N_{i^*=1} I(\Delta^{(i^*)}) 
\end{equation}
where $I(\Delta^{(u^*)})=1 \: if \: \Delta^{(u^*)}\in [-\tau_1,\tau_2]$, else  $I(\Delta^{(u^*)})=0$, $\tau_1>0, \tau_2>0$.

\begin{equation}\label{eq:mae}
MAE=\frac{1}{N}\sum^N_{i^*=1}|\Delta^{(i^*)}|, \: MSE=\frac{1}{N}\sum^N_{i^*=1}(\Delta^{(i^*)})^2
\end{equation}

\begin{equation}\label{eq:mape-1}
 MAPE=\frac{100}{N}\sum^N_{i^*=1}\frac{|\Delta^{(i^*)}|}{R^{(i^*)}}
\end{equation}

A prediction is false positive (FP) if $\Delta^{(i^*)}<-\tau_1$, and false negative (FN) if $\Delta^{(i^*)}>\tau_2$.

\section{Benchmarks on Turbofan Engine Dataset}\label{sec:benchmarks}
We provide a comparison of some approaches for RUL estimation on the engine dataset (\textit{test\_FD001.txt}) below:
\begin{table}[h]
\resizebox{\columnwidth}{!}{
\begin{tabular}{l|c|c|c|c|c|c|c} 
\toprule
\textbf{Approach}&\textbf{S}&\textbf{A}&\textbf{MAE}&\textbf{MSE}&\textbf{MAPE}&\textbf{FPR}&\textbf{FNR}\\ 
\midrule
\textbf{Bayesian-1 \cite{mosallam2014data}}&NR&NR&NR&NR&12&NR&NR\\ 
\textbf{Bayesian-2 \cite{mosallam2015component}}&NR&NR&NR&NR&\textbf{11}&NR&NR\\  
\textbf{ESN-KF \cite{peng2012modified}}&NR&NR&NR&4026&NR&NR&NR\\ 
\textbf{EV-KNN \cite{ramasso2013joint}}&NR&53&NR&NR&NR&36&\textbf{11}\\
\textbf{IBL \cite{khelif2014rul}}&NR&54&NR&NR&NR&18&28\\ 
\textbf{Shapelet \cite{khelif2014rul}}&652&NR&NR&NR&NR&NR&NR\\
\textbf{DeepCNN \cite{babu2016deep}}&1287&NR&NR&340&NR&NR&NR\\ 
\textbf{SOM\tablefootnote{Dataset simulated under similar settings}\cite{macmann2016performing}}&NR&NR&NR&297&NR&NR&NR\\
\textbf{SVR\cite{khelif2017direct}}&449&\textbf{70}&NR&NR&NR&NR&NR\\
\textbf{RULCLIPPER\tablefootnote{Unlike this method which tunes the parameters on the test set to obtain the maximum $S$, we learn the parameters of the model on a validation set and still get similar performance in terms of $S$.} \cite{ramasso2014investigating}}&\textbf{216}*&67&10.0&176&20&56&44\\
\textbf{LR-ED$_2$ (Recon-LR$_2$) \cite{malhotra2016multi}}&256&67&9.9&164&18&\textbf{13}&20\\
\hline
\textbf{Embed-LR$_1$(Proposed)}&\textbf{219}&59&\textbf{9.8}&\textbf{155}&19&14&27\\
\bottomrule
\end{tabular}
}
\caption{\label{tab:aircraftPerf1}Performance of various approaches on Turbofan Engine Data. NR: Not Reported.}
\end{table}

\end{document}